\documentclass{article}
\usepackage{spconf,epsfig}
\usepackage{amsmath,amsfonts}
\usepackage{multirow}
\usepackage{threeparttable}
\usepackage{booktabs}

\usepackage[square,sort&compress,numbers]{natbib}
\usepackage{textcomp}
\usepackage{stfloats}
\usepackage{url}
\usepackage{verbatim}
\usepackage{graphicx}
\usepackage{hyperref}
\usepackage{array}

\usepackage{amsmath}
\usepackage{amsfonts} 
\usepackage{amssymb} 
\usepackage{multirow}
\usepackage{threeparttable}
\usepackage{booktabs}
\usepackage{bbm, dsfont}
\usepackage{bbding}
\usepackage{ifsym}

\usepackage[table]{xcolor}

\title{Time Travelling Pixels: Bitemporal Features Integration with Foundation Model for Remote Sensing Image Change Detection}

\name{Keyan~Chen$^1$, Chengyang~Liu$^1$, Wenyuan~Li$^2$, Zili Liu$^{1,3}$, \textit{ Hao~Chen$^3$, Haotian~Zhang$^1$, Zhengxia~Zou$^1$, Zhenwei~Shi$^{1,*}$} \thanks{$^*$Corresponding author}}
\address{$^1$Beihang University, $^2$University of Hong Kong, $^3$Shanghai AI Laboratory}

\begin{document}
%
\maketitle

\begin{abstract}
Change detection, a prominent research area in remote sensing, is pivotal in observing and analyzing surface transformations. Despite significant advancements achieved through deep learning-based methods, executing high-precision change detection in spatio-temporally complex remote sensing scenarios still presents a substantial challenge. The recent emergence of foundation models, with their powerful universality and generalization capabilities, offers potential solutions. However, bridging the gap of data and tasks remains a significant obstacle. In this paper, we introduce Time Travelling Pixels (TTP), a novel approach that integrates the latent knowledge of the SAM foundation model into change detection. This method effectively addresses the domain shift in general knowledge transfer and the challenge of expressing homogeneous and heterogeneous characteristics of multi-temporal images. The state-of-the-art results obtained on the LEVIR-CD underscore the efficacy of the TTP. The Code is available at \url{https://kychen.me/TTP}.
\end{abstract}
\section{Introduction}
\label{sec:intro}
\vspace{-7pt}

As remote sensing technology for earth observation continues to evolve, remote sensing image change detection has surged to the forefront of research in this field. The primary objective is to analyze the changes of interest within multi-temporal remote sensing products. These changes are typically expressed as pixel-level binary classifications (changed/unchanged).
The dynamic attributes of remote sensing surfaces are influenced not only by natural elements but also by human activities. The precise perception of these changes holds immense significance for the quantitative analysis of land cover alterations. This serves as a potent tool for illustrating macroeconomic trends, human activities, and climate changes. The invaluable application is apparent across various domains, encompassing urban expansion, glacier melting, and the evaluation of economic crop yields \cite{chen2021remote, tang2023wnet, bandara2022transformer, wu2023cstsunet, chen2020spatial}.

High-resolution remote sensing images have emerged as a potent tool for intricate change detection. However, executing robust change detection in complex scenarios remains a formidable challenge \cite{chen2023continuous, chen2022resolution}. Change detection primarily concentrates on ``effective changes" amidst ``non-semantic changes" \cite{tang2023wnet, chen2021remote}. Specifically, non-semantic changes instigated by atmospheric conditions, remote sensors, registration, and semantic changes that are irrelevant to downstream applications (``invalid changes") should be disregarded. This presents considerable obstacles to precise change detection.
Deep learning technology has made significant strides in the realm of change detection. For example, algorithms based on CNN can unveil robust features in changing areas with their strong feature extraction capabilities, achieving impressive performance in a variety of complex scenarios \cite{chen2020spatial, daudt2018fully}. Recently, methods anchored on Transformers have further accelerated the advancement of this field. Transformers can capture long-distance dependencies across the entire image, endowing the model with a global receptive field, and opening up new avenues for tasks like change detection that necessitate high-level semantic knowledge \cite{bandara2022transformer, chen2021remote}.
Despite the remarkable success of these methods, their adaptability in complex and evolving spatiotemporal environments is still a considerable distance from practical application. Furthermore, as the model scale expands, the limited annotated data for change detection significantly curtails the potential of these models. While some strides have been made in self-supervised representation learning and simulated data generation, they still fall short in covering the diversity of remote sensing image scenarios caused by spatiotemporal variability. Nor can they propel the performance of large-parameter models across different scenes \cite{chen2022resolution, chenhao2023continuous}.

The potent universality and adaptability of recent foundational models have been firmly established. These models are trained on vast quantities of data, thereby acquiring generalized knowledge and representations \cite{chen2023rsprompter}. Foundational models in the visual domain, such as CLIP \cite{radford2021learning} and SAM \cite{kirillov2023segment}, have been extensively investigated and utilized by researchers. These models are repositories of a wealth of general knowledge, enabling cross-domain transfer and sharing. This significantly diminishes the need for annotated data for specific tasks.
However, current visual foundational models are primarily designed for natural images, which creates a domain gap when these models are employed for change detection tasks in remote sensing images \cite{chen2023rsprompter}. Moreover, while most visual foundational models excel at comprehending single images, they often fall short in extracting homogeneity and heterogeneity from multiple images, particularly when significant changes occur in the images. This capability is crucial for change detection as it necessitates the model to concentrate solely on ``effective changes".

\begin{figure*}[!t]
\centering
\resizebox{0.9\linewidth}{!}{
\includegraphics[width=\linewidth]{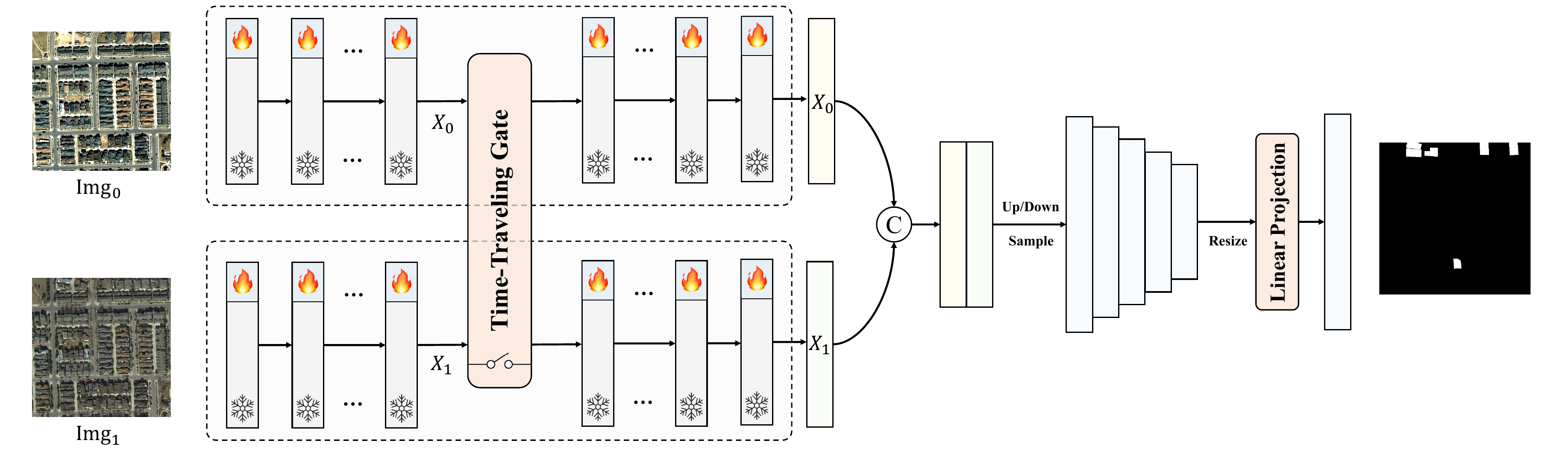}
}
\caption{
The overview of the proposed TTP. The snowflake icon symbolizes that the model parameters are frozen, while the fire signifies training.
}
\label{fig:overview}
\vspace{-14pt}
\end{figure*}

In this paper, we amalgamate the general knowledge of visual foundational models into the task of change detection. This approach overcomes the domain shift encountered during the knowledge transfer and the challenge of expressing the homogeneity and heterogeneity characteristics of multi-temporal images. We introduce Time Travelling Pixels, or TTP, a method that seamlessly integrates temporal information into the pixel semantic feature space.
Specifically, TTP leverages the general segmentation knowledge based on the SAM (Segment Anything) model \cite{kirillov2023segment}. It introduces low-rank fine-tuning parameters into the SAM backbone to mitigate the domain shift of spatial semantics. Furthermore, TTP proposes a time-traveling activation gate that allows temporal features to permeate the pixel semantic space, thereby equipping the foundational model with the capacity to comprehend homogeneity and heterogeneity features between bitemporal images.
Lastly, we devise a lightweight and efficient multi-level change prediction head to decode the dense high-level change semantic features. This innovative approach paves the way for more accurate and efficient change detection in remote sensing images.

The primary contributions of this paper can be encapsulated as follows: 1) We address the issue of insufficient annotated data by transferring the generalized latent knowledge of foundational models to the task of change detection. We introduce the Time Travelling Pixels (TTP) to bridge the time-space domain gap in the knowledge transfer process. 2) More specifically, we incorporate low-rank fine-tuning to mitigate the domain shift of spatial semantics, propose a time-traveling activation gate to augment the foundational model's capacity to discern inter-image correlations and design a lightweight and efficient multi-level prediction head to decode the dense semantic information encapsulated in the foundational model. 3) We compare the proposed method with various advanced methods on the LEVIR-CD dataset. The results demonstrate that our method achieves state-of-the-art performance, underscoring its effectiveness and potential for further applications.

\vspace{-10pt}
\section{Methodology}
\label{sec:method}
\vspace{-10pt}

\subsection{Overview}
\label{sec:overview}
\vspace{-7pt}

To mitigate the annotation requirements of change detection, we leverage the general knowledge transferred from the foundational model. In this paper, we exploit the general segmentation capabilities of the SAM \cite{kirillov2023segment} to construct a change detection network, TTP. TTP is primarily composed of three components: a foundational model backbone based on low-rank fine-tuning; a time-traveling activation gate interposed between dual-temporal features; and an efficient multi-level decoding head. The structure is depicted in Fig. \ref{fig:overview}.


\vspace{-10pt}
\subsection{Efficient Fine-tuning of Foundation Model}
\label{sec:tuning}
\vspace{-5pt}

The backbone of the SAM is comprised of transformer encoders, which can be categorized into base, large, and huge versions, corresponding to 12, 24, and 32 layers, respectively. To bolster computational efficiency, the majority of transformer layers in the backbone employ local attention, with only four layers utilizing global attention. In this study, we leverage the pre-trained, robust visual backbone, maintaining its parameters in a frozen state to expedite adaptation to downstream tasks. To bridge the gap between the domains of natural images and remote sensing images, we introduce low-rank trainable parameters into the multi-head attention layers, as demonstrated in the subsequent equation,
\setlength{\abovedisplayskip}{1pt}
\setlength{\belowdisplayskip}{-10pt}
\begin{align}
\begin{split}
W^* &= W_0 + W_a W_b^T \\
Q = W_q^* X, K &= W_k^* X, V = W_v^* X \\
H &= \text{Softmax}(\frac{Q K^T}{\sqrt{d}}) V \\
\label{eq:tuning}
\end{split} 
\end{align}
where  $W_0 \in \mathbb{R}^{d \times d} $ signifies the original frozen model parameters, while $W_a \in \mathbb{R}^{d \times r}$ and $W_b \in \mathbb{R}^{d \times r}$, $r \ll d$ represent the additional fine-tuning parameters introduced. We incorporate low-rank fine-tuning in the linear projection layer of the self-attention matrix $Q$, $K$, $V$ in each layer of the encoder. $X \in \mathbb{R}^{b \times n \times d} $ denotes the input features, and $H \in \mathbb{R}^{b \times n \times d}$ is the output following the self-attention operation.

 \vspace{-10pt}
\subsection{
Time-traveling Activation Gate
}
\label{sec:gate}
\vspace{-5pt}

Current visual foundational models excel at interpreting the content of single images, yet they fall short in extracting homogenous and heterogeneous features from multiple images. However, in change detection, it is crucial for the model to concentrate on the ``effective differences" in bi-temporal images while disregarding ``irrelevant differences". To tackle this, we introduce the time-traveling activation gate, which facilitates the flow of bi-temporal features into the pixel feature semantic space. This empowers the foundational model to comprehend the changes in bi-temporal images and focus on ``effective changes". For efficiency, we only incorporate the activation gate after the global attention layer in the backbone, \textit{i.e.}, we only employ four bi-temporal time-traveling activation gates. Let's consider $X_0 \in \mathbb{R}^{b \times c \times h \times w }$ and $X_1 \in \mathbb{R}^{b \times c \times h \times w }$ as the features of the previous and subsequent temporal phases, respectively. We follow the formula below to integrate bi-temporal information,
\begin{align}
\begin{split}
M &= \delta(\Phi_{\text{proj}}^1 (\Phi_{\text{cat}}(X_0, X_1))) \\
X_0 &= X_0 + \Phi_{\text{proj}}^2 (M \circ X_1) \\
X_1 &= X_1 + \Phi_{\text{proj}}^2 (M \circ X_0) \\
\label{eq:gate}
\end{split} 
\end{align}
where $\Phi_{\text{cat}}$ symbolizes vector concatenation along the channel dimension, $\Phi_{\text{proj}}^1$ denotes linear channel compression, $\delta$ is a sigmoid activation function, and $\circ$ signifies pixel-wise multiplication. $\Phi_{\text{proj}}^2$ indicates linear mapping.

\vspace{-10pt}
\subsection{
Multi-level Decoding Head
}
\label{sec:decode}
\vspace{-5pt}

Remote sensing image scenes are diverse, and the scale of surfaces can vary significantly. However, visual encoders based on ViT typically generate feature map of a single scale. Despite the map containing high-level global semantic information, their performance advantages can be challenging to demonstrate without multi-level decoding heads. To address this, we introduce a lightweight and efficient multi-level change prediction head. This head constructs multi-level features through transposed convolution upsampling and max pooling downsampling. It then employs a lightweight MLP mapping layer to output the final change probability map, 
\begin{align}
\begin{split}
\{ F_i \} &= \Phi_{\text{sampling}} (\Phi_{\text{cat}}(X_0, X_1)) \\
F_i &= \Phi_{\text{resize}} (\Phi_{\text{proj}}^1 (F_i)) \\
M &= \Phi_{\text{proj}}^2 (\Phi_{\text{cat}} ( \{ F_i \})) \\
\label{eq:decode}
\end{split} 
\end{align}
where $\Phi_{\text{sampling}}$ signifies the feature maps of various levels generated by upsampling/downsampling, $\Phi_{\text{proj}}^1$ and $\Phi_{\text{proj}}^2$ represent the MLP mapping layer, and $\Phi_{\text{resize}}$ refers to applying bilinear interpolation to the features to unify the scale for concatenation.

\vspace{-10pt}
\section{Experiments}
\label{sec:experiment}
\vspace{-7pt}

\subsection{Experimental Dataset and Settings}
\vspace{-5pt}

We carried out experiments on the LEVIR-CD to substantiate the efficacy of our method \cite{chen2020spatial}. This dataset encompasses 637 pairs of bi-temporal images, each with a resolution of $1024 \times 1024$, and includes over 31,333 annotated instances of changes. We adhered to the official standards, partitioning the dataset into three subsets: training, validation, and testing, comprising 445, 64, and 128 image pairs, respectively. 

\vspace{-7pt}
\subsection{Evaluation Protocol and Metrics}
\vspace{-5pt}

To assess the performance, we utilized widely recognized evaluation metrics, including Intersection over Union (IoU), F1 score, Precision, and Recall for the change category, as well as Overall Accuracy (OA) \cite{chen2021remote, bandara2022ddpm}.

\vspace{-7pt}
\subsection{Implementation Details}
\vspace{-5pt}

\textbf{Architecture Details}:
TTP capitalizes on SAM's visual backbone for the transfer of general knowledge. During the low-rank fine-tuning phase, we set $r = 16$. To guarantee efficiency in the decoding head, we limit upsampling to $\frac{1}{4}$ of the original image during supervised training.

\noindent \textbf{Training Details}:
TTP employs a binary cross-entropy function for training. We set the model input size to $512 \times 512$ and utilize data augmentation techniques such as rotation, flipping, random cropping, and photometric distortion to enhance the sample size. During the training phase, the SAM backbone remains frozen. We utilize the AdamW optimizer with a learning rate of 0.0004 and a cosine annealing scheduler with a linear warmup to decay the learning rate. Our batch size is set to 16, and the maximum epoch is 300.

\vspace{-7pt}
\subsection{Comparison with the State-of-the-Art}
\vspace{-7pt}

We have compared the proposed TTP with a series of state-of-the-art change detection methods, including FC-Siam-Di \cite{daudt2018fully}, DTCDSCN \cite{liu2020building}, STANet \cite{chen2020spatial}, SNUNet \cite{fang2021snunet}, BIT \cite{chen2021remote}, ChangeFormer \cite{bandara2022transformer}, ddpm-CD \cite{bandara2022ddpm}, WNet \cite{tang2023wnet}, and CSTSUNet \cite{wu2023cstsunet}. The comparative results are presented in Tab. \ref{tab:levir_sota}. As illustrated in the table, the proposed TTP achieved the highest performance (92.1/85.6 F1/IoU), significantly surpassing the contemporary state-of-the-art methods, WNet (90.7/82.9) and CSTSUNet (90.7/83.0). This underscores that the transfer of general knowledge from the foundational model can bolster the effectiveness of change detection. It also validates the efficacy of the proposed transfer method.

\begin{table}[!htbp] 
\vspace{-10pt}
\centering
\caption{
Comparative results on the LEVIR-CD dataset.
}
\label{tab:levir_sota}
\resizebox{0.98\linewidth}{!}{
\begin{tabular}{c| *{5}{c} }
\toprule
Method 
& $\text{P} $ & $\text{R} $ & $\text{F1} $ 
& $\text{IoU}$ & $\text{OA}$
\\
\midrule
FC-Siam-Di \cite{daudt2018fully} (2018) & 89.5 & 83.3 & 86.3 & 75.9 & 98.7 \\
DTCDSCN \cite{liu2020building} (2020) & 88.5 & 86.8 & 87.7 & 78.1 & 98.8 \\
STANet \cite{chen2020spatial} (2020) & 83.8 & 91.0 & 87.3 & 77.4 & 98.7 \\
SNUNet \cite{fang2021snunet} (2021) & 89.2 & 87.2 & 88.2 & 78.8 & 98.8 \\
BIT \cite{chen2021remote} (2021) & 89.2 & 89.4 & 89.3 & 80.7 & 98.9 \\
ChangeFormer \cite{bandara2022transformer} (2022)& 92.1 & 88.8 & 90.4 & 82.5 & 99.0 \\
ddpm-CD \cite{bandara2022ddpm} (2022)& - & - & 90.9 & 83.4 & 99.1 \\
WNet \cite{tang2023wnet} (2023) & 91.2 & 90.2 & 90.7 & 82.9 & 99.1 \\
CSTSUNet \cite{wu2023cstsunet} (2023) & 92.0 & 89.4 & 90.7 & 83.0 & 99.1 \\
\midrule
TTP (Ours) & 
\cellcolor{gray!30}\textbf{93.0} & \cellcolor{gray!30}\textbf{91.7} & \cellcolor{gray!30}\textbf{92.1} & \cellcolor{gray!30}\textbf{85.6} & \cellcolor{gray!30}\textbf{99.2}
\\
TTP (w/o ttg) & 92.2 & 90.3 & 91.1 & 84.2 & 99.1
\\
TTP (w/o ttg, ml) & 91.9 & 89.3 & 90.6 & 82.8 & 99.0
\\
TTP (w/o ttg, ml, tuning) & 80.9 & 69.3 & 74.6 & 59.5 & 97.6
\\
\bottomrule
\end{tabular}
}
\vspace{-10pt}
\end{table}

\vspace{-7pt}
\subsection{Ablation Study}
\vspace{-7pt}

To thoroughly evaluate the effectiveness of each component, we conducted a series of ablation experiments on the LEVIR-CD dataset, adhering to the same training settings as TTP. As illustrated in Tab. \ref{tab:levir_sota}, the performance experienced a decline when the time travel gate (ttg) and multi-level decoding head (ml) were removed. Moreover, the removal of the low-rank fine-tuning parameters in the foundational model led to a dramatic drop in performance. These observations underscore that the method proposed in this paper can effectively bridge the domain gap and enhance spatio-temporal understanding. They also validate the effectiveness of each component in the change detection task.

\section{Conclusion}
\vspace{-10pt}

In this paper, we tackle the challenge of model generalization in complex spatiotemporal remote sensing scenarios by infusing the generic knowledge of foundational models into the task of change detection. Specifically, we introduce low-rank fine-tuning to bridge the spatial semantic chasm between natural and remote sensing images, thereby mitigating the limitations of the foundational model. We propose a time-travel activation gate to endow the foundational model with the capacity for temporal modeling. Additionally, we design a multi-level change prediction head to decode dense features. Experimental results on the LEVIR-CD dataset underscore the effectiveness of our proposed modules, with the proposed TTP achieving the best performance. This innovative approach paves the way for more accurate and efficient change detection in remote sensing images.

\small{
\bibliographystyle{IEEEbib}
\bibliography{refs}
}

\end{document}